\documentclass[10pt,twocolumn]{ICCAS2019}

\usepackage{graphicx}
\usepackage{microtype}
\usepackage{subfigure}
\usepackage{booktabs} 
\usepackage{amssymb}
\usepackage{amsmath}
\usepackage{xfrac}
\usepackage{nicefrac}
\usepackage{multirow}
\usepackage{multicol}
\usepackage{hyperref}
\usepackage{wrapfig}
\usepackage{xcolor}
\usepackage{array}
\usepackage{gensymb}
\usepackage{pgfplots}

\newcolumntype{P}[1]{>{\centering\arraybackslash}p{#1}}

\begin{document}

\title{Robotic Navigation using Entropy-Based Exploration}

\author{Muhammad Usama and Dong Eui Chang${}^{*}$ }

\affils{ School of Electrical Engineering, KAIST, \\
Daejeon, 34141, Republic of Korea \\
$\{$usama, dechang$\}$@kaist.ac.kr \\ 
${}^{*}$ Corresponding author}


\abstract{
    Robotic navigation concerns the task in which a robot should be able to find a safe and feasible path and traverse between two points in a complex environment. We approach the problem of robotic navigation using reinforcement learning and use deep $Q$-networks to train agents to solve the task of robotic navigation.
    We compare the Entropy-Based Exploration (EBE) with the widely used $\epsilon$-greedy exploration strategy by training agents using both of them in simulation.
    The trained agents are then tested on different versions of the environment to test the generalization ability of the learnt policies. We also implement the learned policies on a real robot in complex real environment without any fine tuning and compare the effectiveness of the above mentioned exploration strategies in the real world setting.
    Video showing experiments on TurtleBot3 platform is available at \url{https://youtu.be/NHT-EiN_4n8}.
}


\maketitle


\section{Introduction}
The presence of various kinds of mobile robots, such as walkers, manipulators etc. is rapidly increasing in the industrial and service sector. Mobile robots have the advantage of simplicity of manufacturing and mobility in the complex environments.
Due to the growing interest of robot utilization in real world environments, the problem of autonomous robotic navigation has garnered increased interest of the research community in recent years. Navigation can be roughly described as the task of finding a feasible path between two points in the surrounding environment \cite{nav-1}.
In robotic navigation, a robot is required to find a collision free and safe path from its current location to some goal location in an unknown, and sometimes dynamic, environment. Since the robot surroundings may contain several static and dynamic obstacles, it is important for the robot to actively seek its goal location while safely avoiding the obstacles and potentially dangerous and undesirable objects, if any. The solution of the complex problem of autonomous robot navigation involves dealing with issues of varied nature, such as acquisition and processing of sensory data, decision making, trajectory generation, trajectory tracking among others.

In this paper, we present the solution to autonomous robotic navigation problem using deep reinforcement learning. We use deep Q-learning to train agents for this task. Moreover, we adopt the entropy-based exploration (EBE) \cite{Usama_EBE}, an exploration strategy based on entropy, that is able to effectively and efficiently explore the state space resulting into better learning than the famous $\epsilon$-greedy exploration heuristic that is widely used among the robotic community. We carry out experiments under diverse conditions to compare both these exploration strategies.


\section{Preliminaries}
\subsection{Reinforcement Learning}

Reinforcement learning is a sequential decision making process in which an agent interacts with an environment $\mathcal{E}$ over discrete time steps; see \cite{suttonAndBarto} for an introduction. While in state $s_{t}$ at time step $t$, the agent chooses an action $a_{t}$ from a discrete set of possible actions i.e. $a_{t} \in {\mathcal{A}}=\{1, \dots , |\mathcal{A}|\}$ following a policy $\pi(s)$ and gets feedback in form of a scalar called reward $r_{t}$ following a scalar reward function, $r:\mathcal{S} \times \mathcal{A} \rightarrow \mathbb{R}$. As a result, the environment transitions into next state $s_{t+1}$ according to transition probability distribution $\mathcal{P}$. We denote $\gamma \in (0,1]$ as discount factor and $\rho_{0}$ as initial state distribution.

The goal of any RL algorithm is to maximize the expected discounted return $R_t = \mathbb{E}_{\pi, \mathcal{P}}$ $[\sum_{\tau=t}^{\infty}\gamma^{\tau-t}r_{\tau}]$ over a policy $\pi$. The policy $\pi$ gives a distribution over actions in a state.

Following a stochastic policy $\pi$, the state dependent action value function and the state value function are defined as
\[
Q^{\pi}(s,a) = {\mathbb{E}} [R_t | s_t = s, a_t = a, \pi ] ,
\]

\[
Q^{\pi}(s,a) = E[R_t | s_t = s, a_t = a, \pi ] ,
\]
and
\[
V^{\pi}(s) = {\mathbb{E}}_{a \sim \pi(s)}[Q^{\pi}(s,a)].
\]
\subsection{Deep $Q$-Networks in Reinforcement Learning}
To approximate high-dimensional action value function given in preceding section, we can use deep $Q$-network (DQN): $Q(s,a;\theta)$ with trainable parameters $\theta$. To train this network, we minimize the expected squared error between the target $y_i^{DQN} = r + \gamma \max_{b}Q(s', b ; \theta^-)$ and the current network prediction $Q(s,a;\theta_i)$ at iteration $i$. The loss function to minimize is given as
\[
L_i(\theta_i) = {\mathbb{E}}[(Q(s,a;\theta_i) - y_i^{DQN})^2],
\]
where $\theta^-$ represents the parameters of a separate \textit{target network} that greatly improves the stability of the algorithm as shown in \cite{humanlevel}. We refer the reader to \cite{deep_neural_mathematical} for formal introduction to deep neural networks.

\begin{table}[t]
    \centering
    \caption{Relation between the agent's action and corresponding angular velocity of the robot.}
    \label{tab:action_to_angular_velocity}
    \begin{tabular}{c|c}
    \toprule
        Action & Angular Velocity of Robot (rad/s) \\
        \midrule
        0 & -1.5 \\
        1 & -0.75 \\
        2 & 0.0 \\
        3 & 0.75 \\
        4 & 1.5 \\
    \bottomrule
    \end{tabular}
\end{table}

\subsection{Entropy}
Let us have a discrete random variable $X$ that is completely defined by the set $\mathcal{X}$ of values that it takes and its probability distribution $\{ p_{X}(x) \}_{x \in \mathcal{X}}$. Here we assume that $\mathcal{X}$ is a finite set, thus the random variable $X$ can only have finite realizations. The value $p_{X}(x)$ is the probability that the random variable takes the value $x$. The probability distribution $p_{X}:X \rightarrow [0,1]$ must satisfy the following condition
\[
\sum_{x \in \mathcal{X}} p_{X}(x) = 1.
\]
The entropy $H_X$ of a discrete random variable $X$ with probability distribution $p_{X}(x)$ is defined as

\begin{align*}
H_X & = - \sum_{x \in {\mathcal{X}}} p_{X}(x) \log_{b}p_{X}(x) \\
    & = - \displaystyle {\mathbb{E}}_{X \sim p_{X}} [\log_{b} p_{X}(x)],
\end{align*}

where the logarithm is taken to the base $b$ and we define by continuity that $0\log_{b}0 = 0$. \\~\
Intuitively, entropy quantifies the uncertainty associated with a random variable. The greater the entropy, the greater is the \textit{surprise} associated with realization of a random variable.

\begin{table*}[t]
\centering
\caption{$Q$- network architecture used to approximate high dimensional $Q$ function. The neteork consists of two fully conencted layers, one dropout layer and an output layer. Here, we have $|{\mathcal{A}}| = 5$.}
\label{tab:qnetwork-architecture}
\begin{sc}
\begin{tabular}{l c c c c c c}
\toprule
Layer & input size & no. of neurons & activation & output size \\
\midrule
FC 1    & $F_{in}$  & $64$  &  RELU  & $64$ \\
FC 2    & $64$  & $64$  &  RELU  & $64$ \\
Dropout, $p_{dropout}=0.2$      & $64$  & $64$  &  -  & $64$ \\
FC 3 (output layer)      & $64$  & $|\mathcal{A}|$  &  linear  & $|\mathcal{A}|$ \\
\bottomrule
\end{tabular}
\end{sc}
\end{table*}

\section{Approach}
In reinforcement learning, an agent is trained to perform a task by maximizing accumulative reward that it gets as a feedback for its interactions with the environment. We use deep $Q$-learning for autonomous robotic navigation where the state observation consists of $360$\degree LiDAR scan and the distance between the current robot position and the desired goal position. The LiDAR scan is generated by means of 360 Laser Distance Sensor (LDS) that is present on the robot. The robot is provided with a constant linear velocity of 0.15m/s in the forward direction. In each state, the agent decides the angular velocity of the robot by choosing an action from a set of five possible actions, ${\mathcal{A}} = \{0,1,2,3,4\}$. Each action $a \in \mathcal{A}$ corresponds to a specific angular velocity of the robot and this correspondence is given in Table \ref{tab:action_to_angular_velocity}. The details about the reward function are given in Section \ref{section:reward}. 

Efficient exploration is crucial for effective learning in reinforcement. The need for good exploration strategy grows ever more in deep reinforcement learning where we have to deal with high dimensional state and action spaces and complex (possibly nonlinear) function approximation architectures. To carry out efficient exploration of the state space, we employ entropy-based exploration strategy that is based on the concept of quantifying the agent's learning in a state based on the difference between the state dependent action values. EBE devises a probability distribution $p_s(a)$ over actions in the state $s$ based on the entropy of state dependent action values $Q(s,a)  \forall a \in \mathcal{A}$ and uses this probability distribution to explore the state space. For the states where some actions are decisively better than other actions, the probability distribution $p_s(a)$ of a learnt policy is highly skewed towards the better actions, thereby, reducing the entropy in that state. Details about EBE exploration strategy are given in Section \ref{sec:EBE_robotics}.

\section{About the Reward Function}
\label{section:reward}

\begin{figure}[t]
    \centering
    \includegraphics[scale=0.325]{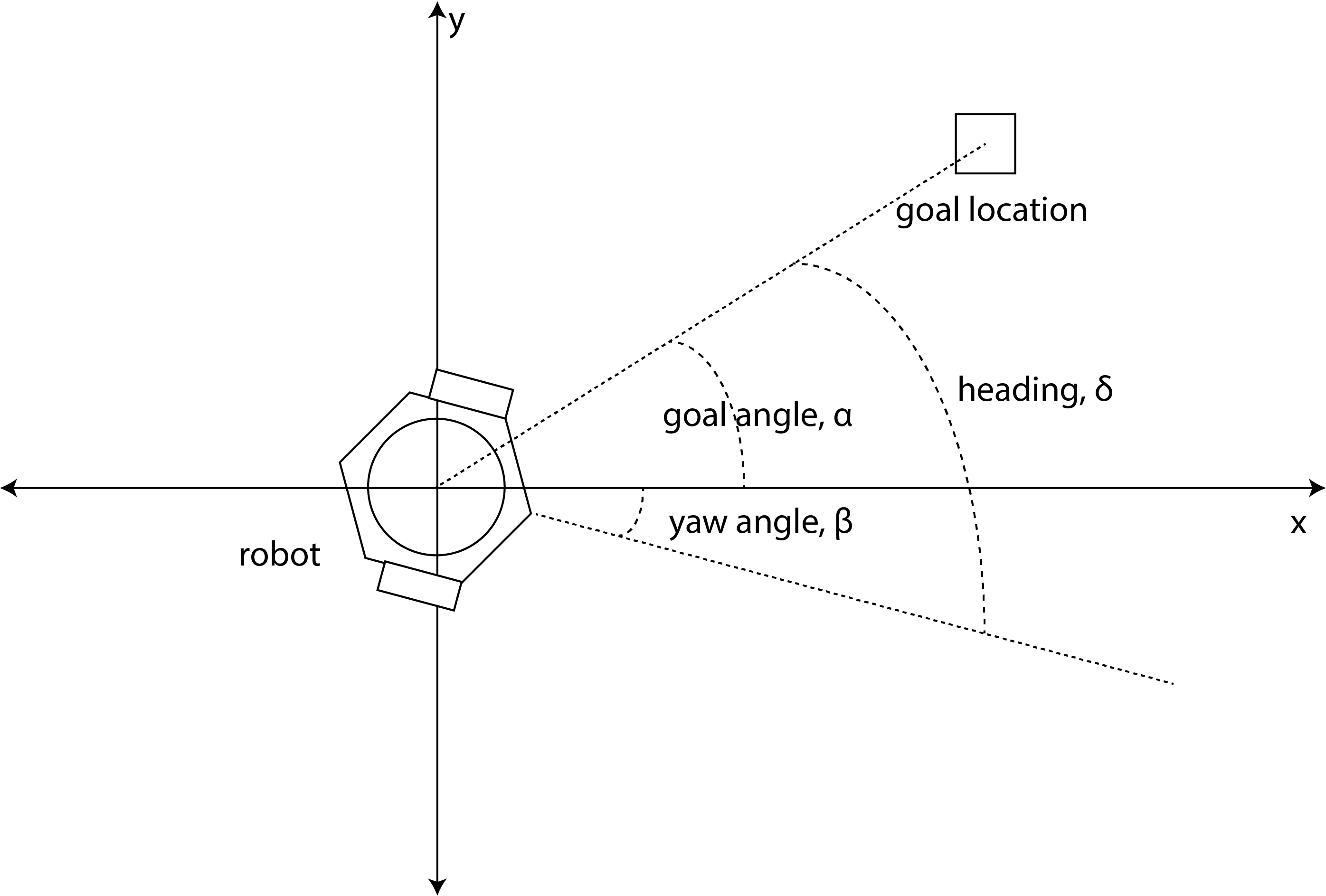}
    \caption{Heading angle $\delta$. $\delta = \alpha - \beta$ where $\alpha$ gives the direction in which robot is travelling and $\beta$ is the yaw angle of the robot. Both $\alpha$ and $\beta$ are defined with repsect to horizontal $+x$ axis.}
    \label{fig:heading}
\end{figure}

The reward function is based on the notion that the agent should receive positive reward for moving towards the goal position and negative reward for moving away the goal position. Therefore, we define the reward function as
\[
    R =
    \begin{cases} 
      R_+ & \text{reaches goal position,} \\
      R_- & \text{collides,} \\
      R_{orientation} * R_{distance} & \text{otherwise,}
   \end{cases}
\]
where $R_{orientation} = 5\psi(\theta)$ and $\theta = -\pi/2 + \delta + a\pi/4$. Here, $a \in \mathcal{A}$ is the action applied by the agent and $\delta$ is the heading angle. Heading angle $\delta$ gives the angle between the direction the robot is travelling and the goal location. It is given by

\[
\delta = \alpha - \beta,
\]
where $\alpha$ is the goal angle between the goal position and the horizontal axis and $\beta$ is the yaw angle of robot with respect to horizontal $+x$ axis. See Figure \ref{fig:heading} for a visual explanation of heading angle $\delta$.

\begin{figure}[t]
    \centering
    \includegraphics[scale=0.35]{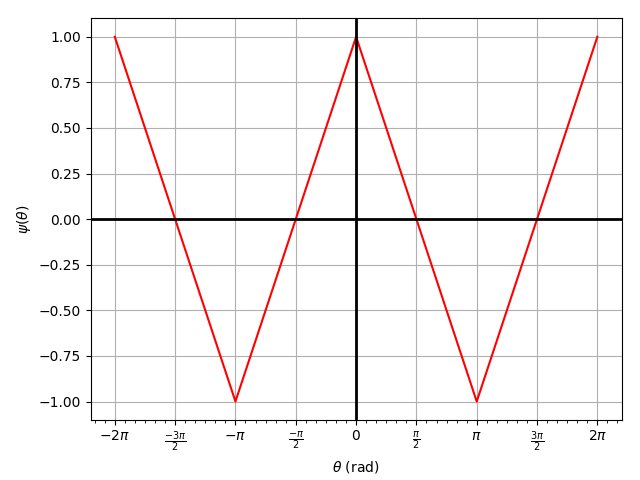}
    \caption{Visualization of function $\psi(\theta)$ for $\theta \in [-2\pi, 2\pi]$}
    \label{fig:reward_function}
\end{figure}

$R_{distance}$ gives the distance reward and is given by $R_{distance} = d_c$ where $d_c$ is the current distance of the robot from the goal location. The function $\psi(\theta)$ is visualized in the Figure \ref{fig:reward_function}. The agent is given a large positive reward $R_+=1000$ for reaching the goal location while it is given a reward of $R_-=-150$ for colliding into an obstacle or a wall.

\section{Entropy-Based Exploration (EBE) for Robotic Navigation}
\label{sec:EBE_robotics}
Entropy-based exploration \cite{Usama_EBE} (EBE) uses the difference between $Q$-values in a state as an estimate of agent's learning progress in that state. Defining a probability distribution over the actions in a state $s$, we have
\begin{equation}
    p_s(a) = \frac{e^{Q(s,a)-Q_{o}(s)}}{\sum_{b \in \mathcal{A}}e^{Q(s,b)-Q_{o}(s)}},
    \label{eq:action_distribution}
\end{equation}
where $Q_{o}(s) = \max_{\Tilde{a} \in {\mathcal{A}}} Q(s,\Tilde{a})$. We then use $p_s(a)$ to obtain state dependent entropy, $\Tilde{H}(s)$, as follows
\begin{equation}
\Tilde{H}(s) = - \sum_{a \in \mathcal{A}} p_{s}(a) \log_{b}p_{s}(a),
\label{eq:EBE}
\end{equation}
where $b > 0$ is the base of logarithm. We note that $\Tilde{H}(s)$ may be greater than $1$ when $|\mathcal{A}| > b$, therefore, we normalize $\Tilde{H}(s)$ between 0 and 1. The maximum value the entropy can take is $\log_{b}(|\mathcal{A}|)$, therefore, we define a scaled entropy $H(s) \in [0,1]$ as follows:
\begin{align}
    H(s) & =   \frac{- \sum_{a \in \mathcal{A}} p_{s}(a) \log_{b}p_{s}(a)}{\log_{b}(|\mathcal{A}|)} \nonumber \\
     & = - \sum_{a \in \mathcal{A}} p_{s}(a) \log_{|\mathcal{A}|}p_{s}(a) 
    \label{eq:EBE_base}.
\end{align}
Here ${\mathcal{A}} = \{0,1,2,3,4\}$, therefore, we have $|{\mathcal{A}}| = 5$.
Given $H(s)$ in a state from equation \eqref{eq:EBE_base}, the agent explores with probability $H(s)$ i.e. it behaves randomly. In practice, entropy-based exploration is similar to the famous $\epsilon$-greedy exploration method with $\epsilon$ replaced with state dependent $H(s)$.

\section{Experiments}
In this section, we share details about various experiments performed on the Turtlebot3 platform. Experiments were performed in different environment configurations as detailed below.

We use deep neural network to approximate the high dimensional state dependent action value function. The architecture of deep neural network used is given in Table \ref{tab:qnetwork-architecture}.

\subsection{Environments with no obstacles}
\label{sec:no_obstacles}
In this configuration, the environment consists of square maze in which the robot is tasked to the reach randomly generated goal positions. The episode ends when the robot collides with either of the maze walls or it has consumed 500 time steps. The task of the agent is to go to randomly generated goal positions without colliding with the maze walls. The reward function is the same as described in Section \ref{section:reward}.

We compare the Entropy-Based Exploration (EBE) strategy with the famous $\epsilon$-greedy exploration heuristic. For the baseline $\epsilon$-greedy exploration heuristic, the exploration fraction in each episode, $\epsilon$, is given for episode $i$ as
\[
    \epsilon_i = max(\alpha \epsilon_{i-1} , \beta)
\]
for $i = 2, \dots, N$ where $N$ is the total number of episodes. Here, $\alpha = 0.99$, $\epsilon_1 = 1.0$ and $\beta = 0.05$.

\begin{figure}[t]
    \centering
    \subfigure[4x4 square maze on which the training is performed.]
    {
        \includegraphics[scale=.35]{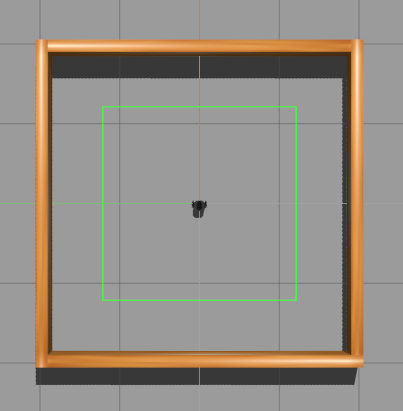}
    }
    \\
    \subfigure[Training progress comparing EBE and $\epsilon$-greedy exploration strategies.]
    {
        \includegraphics[scale=.375]{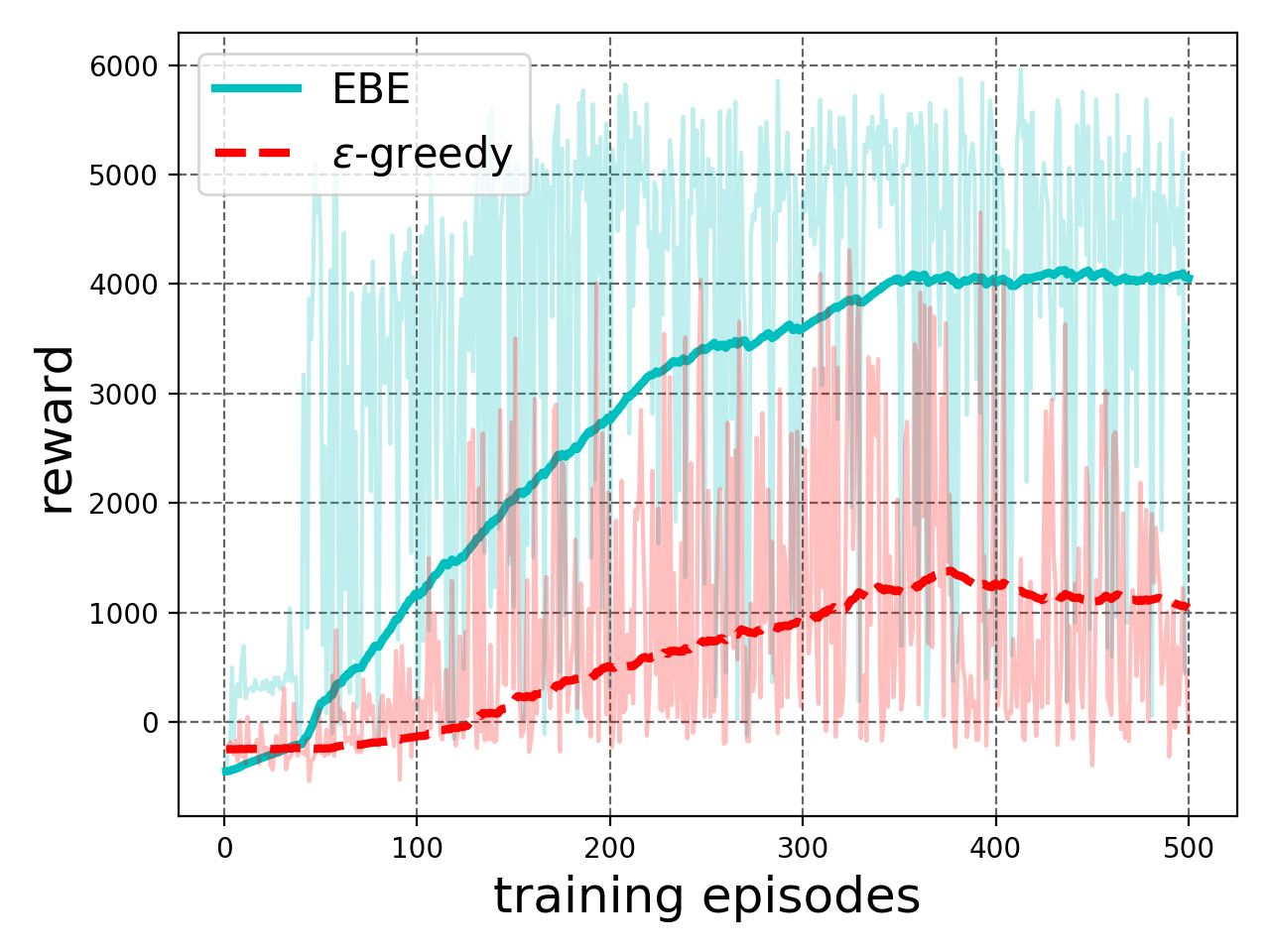}
    }
    \caption
    {
    For environment with no obstacles, training is performed on 4x4 maze using EBE and $\epsilon$-greedy exploration strategies.
    }
    \label{fig:no_obstacle}
\end{figure}

The agents are trained on a 4x4 square maze shown in Figure \ref{fig:no_obstacle}(a) and the training progress is shown in Figure \ref{fig:no_obstacle}(b). Both agents are trained using deep $Q$-learning algorithm for 500 episodes. The details about the hyperparameters used in the training process is given in Table \ref{tab:hyperparameetrs}.

\begin{table}[t]
    \centering
    \caption{Hyperparameters used in the experiments.}
    \begin{tabular}{c|c}
    \toprule
        Hyperparameter & Value \\
    \midrule
         maximum episodes & 500 \\
         maximum episode steps & 500 \\
         discount factor & 0.9 \\
         learning rate & 0.00025 \\
         target network update frequency (in steps) & 2000 \\
         batch size & 64 \\
         replay memory size & $10^6$ \\
     \bottomrule
    \end{tabular}
    \label{tab:hyperparameetrs}
\end{table}

We see in Figure \ref{fig:no_obstacle}(b) that EBE strategy shows better performance both in terms of learning high reward policy as well as learning speed depicting efficient exploration. The policy learnt using $\epsilon$-greedy exploration strategy settles at much lower score.

\begin{table}[b]
    \caption{Mean and standard deviation of rewards obtained in 10 consecutive test episodes of test experiments with environments having no obstacles.}
    \label{tab:no_obstacles}
    \centering
    \begin{tabular}{l | c | c}
    \toprule
        Environment &  EBE & $\epsilon$-greedy \\
        \midrule
        4x4 Square & \textbf{3731.92 $\pm$ 368.35} & 2202.93 $\pm$ 425.89 \\
        8x8 Square & \textbf{2638.52 $\pm$ 319.07} & 1662.57 $\pm$ 402.94 \\
        2x2 Square & 4369.64 $\pm$ 935.56 & \textbf{5556.55 $\pm$ 377.02} \\
        \midrule
        Triangle & \textbf{3655.37 $\pm$ 915.96} & 1678.71 $\pm$ 1257.50 \\
        Pentagon & \textbf{2794.04 $\pm$ 544.07} & 2065.00 $\pm$ 882.77 \\
    \bottomrule
    \end{tabular}
\end{table}

\begin{figure}[t]
    \centering
    \subfigure[triangular maze]
    {
        \includegraphics[scale=.2]{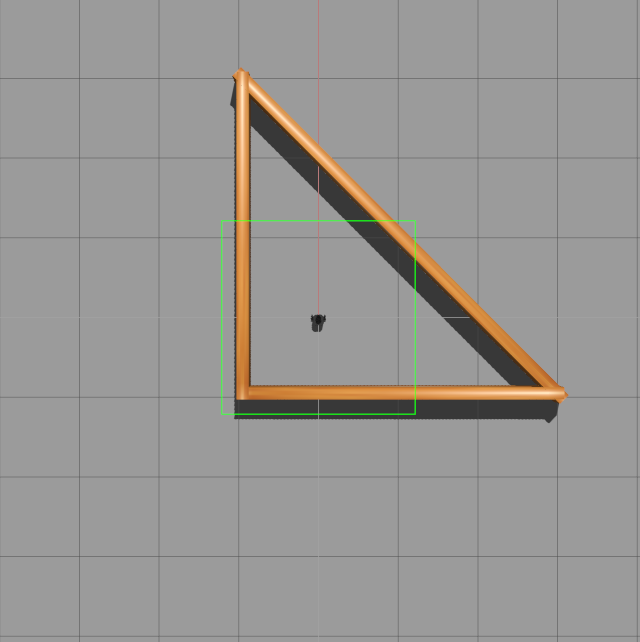}
    }
    \\
    \subfigure[pentagonal maze]
    {
        \includegraphics[scale=.2]{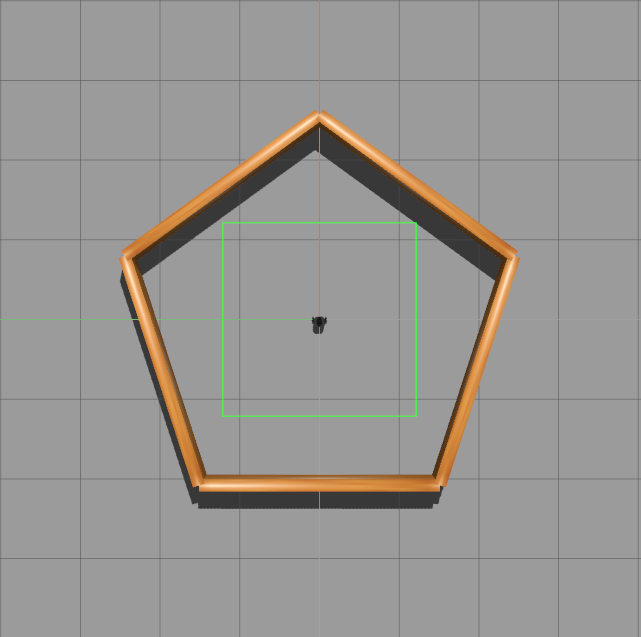}
    }
    \caption
    {
        Agents trained on 4x4 square maze (Figure \ref{fig:no_obstacle}(a)) are tested additionally on (a) triangular maze and (b) pentagonal maze to test the robustness and generalization ability of the learnt policies.
    }
    \label{fig:tri_penta}
\end{figure}

To test the robustness and generalization ability of the learnt policies, we test the agents trained on 4x4 square maze on 2x2, 4x4 and 8x8 square mazes. Also, we test both these agents on mazes of different shapes including a triangular maze shown in Figure \ref{fig:tri_penta}(a) and a pentagonal maze shown in Figure \ref{fig:tri_penta}(b). Note that mazes shown in Figure \ref{fig:tri_penta} were never shown to the agent during the training process.
The results of these test experiments are reported in Table \ref{tab:no_obstacles}. We see in Table \ref{tab:no_obstacles} that policy learnt using EBE exploration shows better performance on 4x4 and 8x8 square mazes. It, however, shows lags behind the policy learnt using $\epsilon$-greedy policy on 2x2 square maze. The low score by EBE policy on 2x2 maze can be explained by noting that the EBE policy learns to take long radius turns as compared to the $\epsilon$-greedy policy. Since 2x2 maze is essentially smaller in size as compared to 4x4 and 8x8 mazes, for the goal positions located in close proximity of the maze walls, they possibility of collision with the walls is greater for EBE policy as compared $\epsilon$-greedy policy, resulting in low score for EBE policy. Table \ref{tab:no_obstacles} also shows the results of experiments with triangle and pentagonal mazes where the EBE policy obtains better scores than the $\epsilon$-greedy policy. Note that these agents were trained only on the 4x4 maze shown in Figure \ref{fig:no_obstacle}(a).

\begin{figure*}[t]
    \centering
    \subfigure[]
    {
        \includegraphics[scale=.3525]{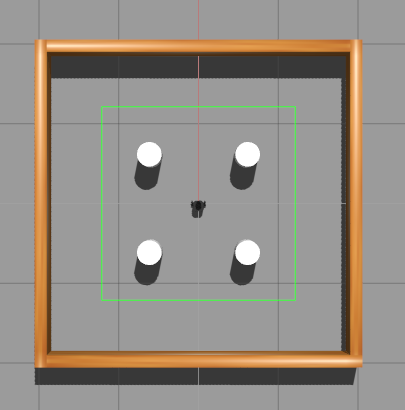}
    }
    \subfigure[]
    {
        \includegraphics[scale=.3]{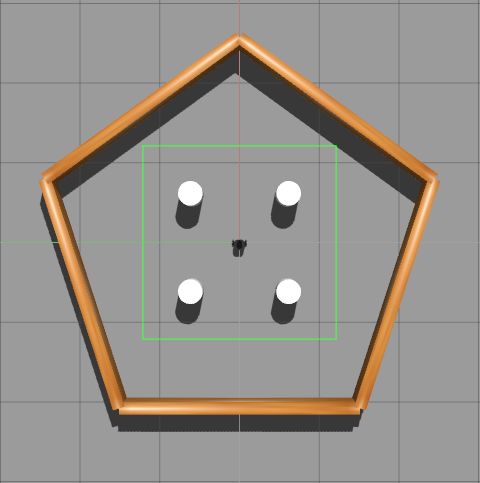}
    }
    \subfigure[]
    {
        \includegraphics[scale=.4]{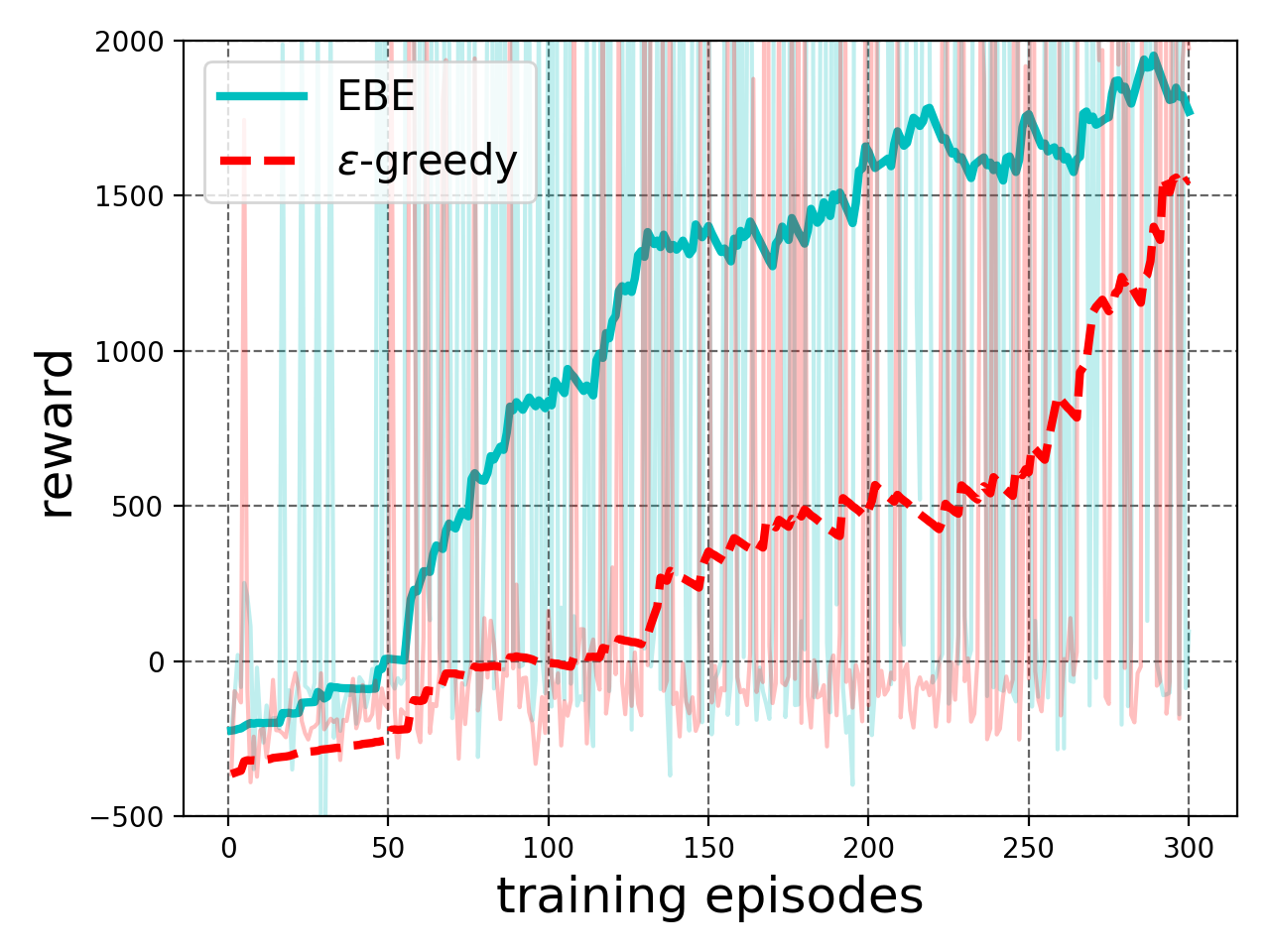}
    }
    \caption
    {
        (a) 4x4 square maze with obstacles on which the training is performed.
        (b) Pentagonal maze with obstacles used in test experiment.
    (c) The training process comparing the EBE exploration strategy against $\epsilon$-greedy exploration heuristic.
    }
    \label{fig:static_obstacles}
\end{figure*}

\begin{table}[t]
    \caption{Mean and standard deviation of rewards obtained in 10 consecutive test episodes of test experiments with environments having no obstacles.}
    \label{tab:static_obstacles}
    \centering
    \begin{tabular}{l | c | c}
    \toprule
        Environment &  EBE & $\epsilon$-greedy \\
        \midrule
        4x4 Square & \textbf{1125.45 $\pm$ 775.82} & 450.19 $\pm$ 760.14 \\
        Pentagon & \textbf{890.99 $\pm$ 641.19} & 432.28 $\pm$ 210.37 \\
    \bottomrule
    \end{tabular}
\end{table}

\subsection{Environments having obstacles}
\label{sec:static-obstacles}
Here, the robot again has to reach randomly generated goal positions but in the presence of additional obstacles. These are static obstacles and the episode ends when the robot collides with either of these static obstacles or any of the maze walls.

The training conditions, hyperparameters, baseline $\epsilon$-greedy exploration schedule and reward function are same as explained in Section \ref{sec:no_obstacles}. The training environment is shown in Figure \ref{fig:static_obstacles}(a) while the training progress comparing the EBE and $\epsilon$-greedy exploration strategies are shown in Figure \ref{fig:static_obstacles}(c). It can be seen in Figure \ref{fig:static_obstacles}(b) the EBE shows better learning in terms of learnt policy and learning speed. Both policies are tested on a 4x4 square maze (Figure \ref{fig:static_obstacles}(a)) and a pentagonal maze (Figure \ref{fig:static_obstacles}(b)).

The experiment results are given in Table \ref{tab:static_obstacles}. We see that EBE exploration strategy outperforms the $\epsilon$-greedy exploration strategy on both environments with considerate margin depicting effective exploration.

\section{Conclusion}
We consider the problem of autonomous navigation for robotics and apply deep reinforcement learning to solve this problem. We use entropy-based exploration strategy and the widely famous $\epsilon$-greedy exploration heuristic to train agents using deep $Q$-learning algorithm to solve achieve the autonomous robotic navigation. We performed experiments under various environmental conditions to test the effectiveness of both of the above mentioned exploration strategies.

\section{Acknowledgement}
This research has been in part supported by the ICT R\&D program of MSIP/IITP [2016-0-00563, Research on Adaptive Machine Learning Technology Development for Intelligent Autonomous Digital Companion].





\end{document}